# The Complexity of Reasoning about Spatial Congruence

**Matteo Cristani**                                         CRISTANI@SCI.UNIVR.IT
*Dipartimento Scientifico e Tecnologico,*
*Università di Verona,*
*Cà Vignal 2, strada Le Grazie, I-37134 Verona*

## Abstract

In the recent literature of Artificial Intelligence, an intensive research effort has been spent, for various algebras of qualitative relations used in the representation of temporal and spatial knowledge, on the problem of classifying the computational complexity of reasoning problems for subsets of algebras. The main purpose of these researches is to describe a restricted set of maximal tractable subalgebras, ideally in an exhaustive fashion with respect to the hosting algebras.

In this paper we introduce a novel algebra for reasoning about Spatial Congruence, show that the satisfiability problem in the spatial algebra MC-4 is NP-complete, and present a complete classification of tractability in the algebra, based on the individuation of three maximal tractable subclasses, one containing the basic relations. The three algebras are formed by 14, 10 and 9 relations out of 16 which form the full algebra.

## 1. Introduction

Qualitative spatial reasoning has received an increasing amount of interest in the recent literature. The main reason for this, as already observed by Jonsson and Drakengren (1997), is probably that spatial reasoning has proved to be applicable to real-world problems, as in Geographical Information Systems (Egenhofer, 1991; Grigni, Papadias, & Papadimitriou, 1995), and Molecular Biology (Cui, 1994).

The specific stress on *qualitative* reasoning about space, as observed by Zimmermann (1995), is justified by the fact that qualitative spatial relations can be treated as efficiently as their quantitative counterparts, but they seem to be closer to the model of relations humans adopt for spatial reasoning.

Even though qualitative spatial reasoning has an extended literature, in spite of its relatively short history, certain aspects of the discipline have been neglected. In particular, no exhaustive computational perspective has been developed on *qualitative morphological reasoning about space*. The term morphological reasoning is intended to suggest reasoning about the *internal structure* of the objects. In the case of spatial reasoning this includes reasoning about the size, shape and internal topology of spatial regions.

The purpose of the present work is to analyse the complexity of reasoning about relations of congruence, either actual or partial, between spatial regions, using the spatial algebra MC-4 which has been preliminarly analysed by Cristani (1997).

The algebra MC-4 is a Constraint Algebra (Ladkin & Maddux, 1994) for qualitative reasoning about the morphological relation of congruence. Two spatial regions, in the models documented in literature, are considered to be equivalent if and only if they share interior and boundaries, namely if and only if they are the *same* spatial region. In particular, the relation EQ, as defined by Randell, Cui and Cohn (1989), and analogously by Egenhofer and Franzosa (1991), becomes *identity* under the *unique name assumption*. Geometry and topology, conversely, allow other kinds of equivalence relations. These relations are weaker than the identity, namely the equivalence classes they induce are larger than singletons.

The simplest weakening we can define is the *congruence* relation. Even though, the relation has been studied (Borgo, Guarino, & Masolo, 1996, 1997; Cristani, 1997; Mutinelli, 1998), the complexity of reasoning in subalgebras has not yet been deeply investigated.

The reason to introduce this new relation, and to provide an algebraic structure to host relations based on it (in particular relations in which we compare regions, which, even if they are not congruent, can be overlapped by roto-translating one into the interior of the other), is that, in many cases, the knowledge we have to represent in our systems necessarily includes internal structures. Consider the following example.

**Example 1**

A GIS is dedicated to the representation of geographical data about industrial sites.

In the system, the structure of factories can be described by means of various attributes, including size and shape. One of the users wants to query the system about the opportunity of moving the factory, where he works, from the present site to a new one. In the new site prefabricated facilities already exist, and the problem consists of deciding how to preserve topological layout of the factory and minimise the costs of buying new engines, to substitute those which cannot fit with the new facilities.

In order to reason about these problems the system can be made able to represent the qualitative relations between the old and new site parts, establishing, in particular whether the spatial regions occupied by the engines are "congruent", or even "congruent to a part" of the ones which can be chosen for hosting them in the new factory's site.

Clearly we cannot use topological information, since a spatial region presently occupied by an engine is surely disjoint from the region that this engine is going to occupy in the new site, but this information is insufficient for deciding whether the new site is able to host the engine itself.

Though metric information can be involved in final decision about moving the factory, qualitative reasoning can be used in the initial design phase.

In Figure 1 we give a pictorial representation of one possible situation in which, to make decisions, we need to represent spatial information which is not topological.

□

The algebra MC-4 can be used to represent the four basic relations which can be built on the equivalence relation of congruence: if we consider the roto-translation of a region $x$ with respect to a region $y$, only four possible situations arise.

- There is at least one roto-translation in which $x$ coincides with $y$. The two regions are *congruent*.

- There is at least one roto-translation in which $x$ coincides with a proper part of $y$. The region $x$ is *congruent to a proper part of $y$*.

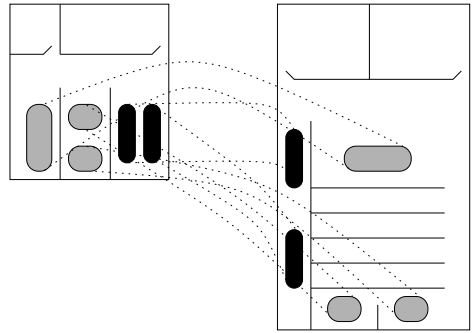

Figure 1: A pictorial representation of the relations to be analysed in a GIS for moving a factory from an old to a new site. Note that the production lines of the new facility may not be suitable for hosting certain engines.

- There is at least one roto-translation in which $y$ coincides with a proper part of $x$. The region $x$ *has a proper part congruent to* $y$.

- None of the above. The region $x$ and $y$ *cannot be perfectly overlapped.*

We would like to stress here three main aspects of these relations.

1. The relations can be established between two spatial regions in any position of space. There is no specific need for the regions to be close, or to be in any particular topological relation, provided that the relation is compatible; so these congruence relations are purely *morphological.*

2. The equality of two regions implies congruence. The relation of proper part excludes morphological relations other than "congruent to a part". This is not the case for the morphological side. A region can be congruent to a part of another region, even if they are disjoint, and the same can happen to congruence, as already stated in point 1.

3. Even if two regions are of the same size, there may be the case that the regions are not congruent. This holds also for partial congruence, being possible that a region is smaller than another one but they cannot be overlapped.

The practical relevance of the algebra can be proved by exhibiting many other examples, especially coming from GIS. In this paper we deal with the problem of reasoning about congruence in two-dimensional space domain. The kind of congruence we assume is the weakest one: roto-translability. Three-dimensional congruence is not of the same type. The "natural" notion of three-dimensional congruence is isometry, which is much more general than roto-translability. For example left and right hands are congruent (under the simplified assumption that they have the same shape and size), but they cannot be roto-translated each into the other one. We concentrate here in two-dimensional reasoning which may be analysed in terms of roto-translation and topological relations. The analysis of isometries is left for further work.

The paper is organised as follows. Section 2 describes related work which has been developed in the area. Section 3 presents the spatial algebra MC-4, and Section 4 discusses the classification of tractability we found for this algebra. Finally, in Section 5, some conclusions are given.

## 2. Previous Work

From the perspective of individuating primitives to describe space, a significant effort has been lately spent, in the direction of defining *binary* relations between spatial regions, which may be used as a model of space in *qualitative* terms. Moreover, it was natural, in the AI community, to use Constraint Processing for reasoning about such binary relations. A specific attention to Constraint Processing emerged in the Spatial Database community, and in the community of Geographical Information Science as well.

Two apparently independently developed models, which can be shown to be essentially equivalent, exist in the Artificial Intelligence (Randell & Cohn, 1989; Randell, Cui, & Cohn, 1992) and Spatial Database literatures (Egenhofer & Herring, 1990; Egenhofer & Franzosa, 1991; Franzosa & Egenhofer, 1992). The Artificial Intelligence model, (Randell & Cohn, 1989; Randell et al., 1992) is known as RCC, where the acronym stands for **Region Connection Calculus**.

The RCC model is centered on the primitive of "connection" as originally suggested by Clarke (1981). Gotts (1994) and further Gotts, Gooday and Cohn (1996) obtained connection to be the only primitive. In the original framework (Randell & Cohn, 1989; Randell et al., 1992) the model was formulated in two versions, now called RCC-5 and RCC-8. The RCC-5 model is a model in which we cannot distinguish between interior and boundary of a spatial region (so that the external connection may be tangency or overlapping, for example), while in the RCC-8 model this distinction is possible.

Bennett (1994, 1995), used propositional logic to represent RCC reasoning problems. He observed that, given a propositional logic, and interpreting the truth values of each formula as a spatial region, the language of RCC-5 is sufficient to express a satisfaction problem at the semantical level. This fact, however, is not due to the spatial interpretation, because a non-spatial interpretation in which the RCC-5 relations are still sufficient to represent truth condition at semantical level can be found: set theory. Therefore, he applied the result of Cook (1971) on NP-completeness of classical propositional logic to the restricted model RCC-5, proving that deciding the consistency of a Constraint Satisfaction Problem (CSP) on RCC-5 is an NP-complete problem too.

The complete model of RCC-8, instead, can not be analysed in terms of pure set theory, because the distinction between boundary-connection and interior-connection is possible. This distinction means that the minimal interpretation in which RCC-8 still provides a correct and complete representation of truth conditions at semantical level requires topology. Statman (1979) proved that intuitionistic logic along with interpretations in set theory forces a topology in the models of a sound theory. Bennett proved, thus, that RCC-8 can be used to define the truth conditions of formulas in an intuitionistic propositional logic. Statman (1979) proved that intuitionistic propositional logic is PSPACE-complete. However, since we do not need a complete truth verification procedure, but only a procedure for constraint processing, Bennett could reduce the result, proving that also RCC-8 is NP-complete.

These results, even if they are in general negative for practical purposes, encouraged researches in the direction of restricted models, in such a way that at least for certain cases we may process a finite set of RCC constraints in polynomial fashion on deterministic machines. In particular, Nebel (1995) showed that reasoning with the basic relations of RCC-5 and RCC-8 are tractable problems. Renz and Nebel (1999) improved the results above, by showing that there exists a maximal tractable subclass of RCC-5, denoted by $\widehat{\mathcal{H}}_5$, formed by 28 relations out of 32, which includes all the basic relations, and a maximal tractable subclass of RCC-8, denoted by $\widehat{\mathcal{H}}_8$, formed by 148 relations out of 256 including the basic relations. A *maximal tractable subclass* $A$, is a subset of a constraint algebra, such that problems defined on $A$ are tractable, while problems defined on proper supersets of $A$ are not. These results have been obtained in a similar fashion to the work of Nebel and Bürckert (1995) on temporal reasoning. The result of Renz and Nebel is anyway incomplete, since he simply proved that there exists one maximal tractable subclass, and he did not classify *every* maximal tractable subclasses of RCC-5. Jonsson and Drakengren (1997) showed that there exist four maximal tractable subclasses of RCC-5, one including the basic relations. The result is obtained in a similar fashion to (Drakengren & Jonsson, 1997; Jonsson, Drakengren, & Bäckström, 1999). A complete analysis of the RCC-8 maximal tractable subclasses including the basic relations has been provided by Renz in (1999).

Our result is the analogous in MC-4 of Jonsson and Drakengren result for RCC-5. The MC-4 algebra, we describe in this paper, is structured in the same way as the Algebra of Partially Ordered Time (PO-time algebra), studied by Anger, Mitra and Rodriguez. We would like to stress two main aspects here:

- The MC-4 algebra has the PO-time algebra are the *same* algebraic structure: the discussion of this paper stands a different interpretation of the PO-time algebra, even different from the spatial interpretation provided by Anger, Mitra and Rodriguez in their papers.

- The computational results we present here can be applied to the PO-time algebra as well, and they extend the previous findings obtained by Anger, Mitra and Rodriguez.

In particular Anger, Mitra and Rodriguez (1998, 1999), proved that path-consistency is insufficient to ensure consistency for relations in this algebra, and that there exists a tractable subalgebra of it which can be treated by an $O(n^3)$ algorithm.

Anger, Mitra and Rodriguez (1999), showed that deciding the consistency of a PO-time network is an NP-complete problem by reducing to it the analogous decision problem on RCC-5.

Some important observations are needed, with respect to the results of Anger, Mitra and Rodriguez:

- The results on the complexity of PO-time algebra can be applied to MC-4, only if we can show that when a MC-4 network is consistent then we can exhibit a consistent scenario in which vertices of the network are substituted with spatial regions. Therefore, even if we can show this correspondence for PO-time algebra too, our NP-completeness result is independent. Moreover, the method we used to prove the intractability of MC-4 is independent as well, and the main difference is in the way we used to exhibit problematic algebraic structure. Through this proof we derived a simple way to solve the problems whenever possible in polynomial time.

- The discovery of the tractable subalgebras we indicate by $M_{99}$, $M_{81}$ and $M_{72}$ in the present paper, deserves acknowledgement of priority to Anger, Mitra and Rodriguez. However, Anger, Mitra and Rodriguez also classified one more algebra, which is tractable, but not maximal, since it is contained in $M_{99}$. Moreover, the algorithms we found for $M_{99}$ and $M_{81}$ are substantially different from the one Anger, Mitra and Rodriguez present and more efficient, being $O(n^2)$ instead of $O(n^3)$. Therefore the two algorithmic solutions can be considered as independent results as well.

- The classification presented here is complete, since we classify *all* the maximal tractable subalgebras of MC-4, and this is a result which may be applied to temporal reasoning as well, since it is obtained by means of algorithms which are completely independent from the interpretation we give to the relations (either spatial or temporal).

- The introduction of *morphological* relations in spatial reasoning is novel too, and the study of spatial congruence deserves, in our opinion, deep investigations henceforth. The fact that two algebras of spatial and temporal reasoning present substantial similarities is not a novelty. The RCC model corresponds to subalgebras of Interval Calculus, as stated by Bennett (1994). Also Anger, Mitra and Rodriguez stated this property of PO-time algebra with respect to RCC-5, but their interpretation is completely different from our own, the equality of PO-time being interpreted as EQ in RCC-5, while we interpret it as congruence.

Thus, even though a similar algebraic structure has been partially investigated before, the present paper presents substantial methodological differences and we have obtained results which are independent of or extensions to the ones obtained by Anger, Mitra and Rodriguez.

In the remainder of the paper, when a result can be attributed to Anger, Mitra and Rodriguez we note it in the text.

## 3. The Spatial Algebra MC-4

In this section we present the spatial algebra MC-4, which has been previously presented by Cristani (1997) and largely analysed in by Mutinelli (1998).

MC-4 is a *Binary Constraint Algebra* (henceforth indicated as a Constraint Algebra). In a Constraint Algebra we have a *Constraint Domain* and an *Algebra Base*, which is formed by mutually exclusive relations among elements of the Constraint Domain, whose union form the universal relation. The converse of a basic relation is a basic relation too, and the composition of basic relations is the union of some basic relations. A *Constraint* is the establishment of one of all possible unions of basic relations between two variables which vary on the domain. A constraint is satisfied by an assignment of one pair of values of the domain to the variables, so that the pair of values is in one of the relations of the constraint itself. Given a finite set of constraints between two variables, the problem of deciding whether there exists an assignment to the variables such that all the constraints can be simultaneously satisfied is referred to as *Constraint Satisfaction Problem*, and henceforth indicated as a CSP.

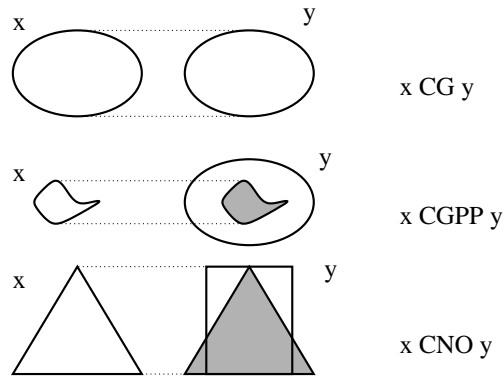

Figure 2: A pictorial representation of the basic relations of MC-4

| $\otimes$ | CG | CGPP | CGPP$^{-1}$ | CNO |
|---|---|---|---|---|
| **CG** | CG | CGPP | CGPP$^{-1}$ | CNO |
| **CGPP** | CGPP | CGPP | $\top$ | CGPP<br>CNO |
| **CGPP$^{-1}$** | CGPP$^{-1}$ | $\top$ | CGPP$^{-1}$ | CGPP$^{-1}$<br>CNO |
| **CNO** | CNO | CGPP<br>CNO | CGPP$^{-1}$<br>CNO | $\top$ |

Table 1: The composition table of MC-4. The symbol $\top$ represents the universal relation $\{\mathsf{CG}, \mathsf{CGPP}, \mathsf{CGPP}^{-1}, \mathsf{CNO}\}$.

The MC-4 algebra is formed by all the unions of the four basic relations, which can be established between two-dimensional spatial regions, from a morphological point of view, with respect to the equivalence relation of *congruence*.

The congruence relation is variously interpreted in geometry. Our interpretation is the most restrictive one: two regions are congruent *iff* they can be rigidly roto-translated into each other.

A region $x$, in this interpretation, may be *congruent* ($x$ CG $y$) to a region $y$, or *congruent to a part of $y$* ($x$ CGPP $y$) or they *cannot be perfectly overlapped* ($x$ CNO $y$). The relation *congruent to a part of* may also be inverted to *having a part congruent to* ($y$ CGPP$^{-1}$ $x$ *iff* $x$ CGPP $y$).

In Figure 2 a pictorial representation of the three basic relations CG, CGPP and CNO is given. In Table 1 we present the composition table of MC-4 showing how the basic relations compose with each other.

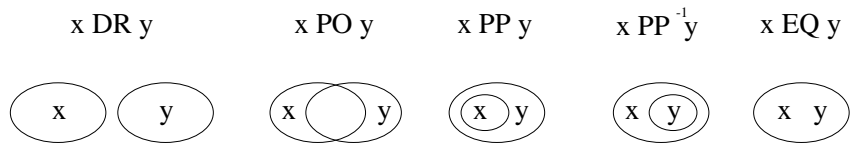

Figure 3: A pictorial representation of the relations of RCC-5.

A CSP can be represented by a *Network of Constraints*. A network of constraints is a labelled graph $\mathcal{G} = \langle \mathcal{V}, \lambda, \mathcal{E} \rangle$, where $V$ is a finite set of *vertices*, $E$ is a binary relation on $V$ whose elements are called *edges*, and $\lambda$ is a *labelling function* which maps each vertex of $\mathcal{G}$ to a variable, and each edge to a relation of a Constraint Algebra on a given domain $D$. Given a network of constraints $\mathcal{G}$, the problem of deciding the consistency of $\mathcal{G}$ is the problem of establishing whether is it possible to instantiate each vertex label (the variables) with elements of $D$, in such a way that all the relations represented in $\mathcal{G}$ as labels of the edges are simultaneously satisfied. The CSP represented by a network $\mathcal{G}$ is often referred to, for the algebra $A$, as A-SAT$(\mathcal{G})$. For the sake of simplicity we refer to the CSP represented by a network $\mathcal{G}$ on MC-4 as MSAT$(\mathcal{G})$.

The main result on the MC-4 algebra, with respect to MSAT, is unfortunately a negative computational account. In general, deciding the consistency is hard to solve for networks of constraints between variables representing spatial regions, as stated in Theorem 2. This result has been already proved by Cristani (1997).

Anger, Mitra and Rodriguez (1998), showed that path-consistency cannot be applied successfully to the Algebra of Partially Ordered Time, which is isomorphic to MC-4 at the syntactic level. This is insufficient to prove that the CSP on this algebra is NP-complete. They showed (Anger et al., 1999) that PO-time algebra is NP-complete. Their proof lies on the $\sigma$ translation, which can be shown to be analogous, but not identical, to the $\Omega$ translation we introduced here, at the syntactical level, being completely different at the semantical one. However, their proof is insufficient to show that we can arrange spatial solutions, since the map they defined is purely syntactical. As already observed by Lemon (1996), the representation of space by means of relation algebra is not pure, and we can obtain satisfiable networks which are not realizable in space. Our proof, instead, can be applied to spatial interpretations. It can be applied to nonlinear time temporal interpretations as well, since the syntactic level is shown to be sufficient for space, and nonlinear time can be interpreted as space too (Anger et al., 1999).

Before getting into the proof of this negative result we need to describe a relevant correspondences of the MC-4 algebra to the RCC-5 model, which is also used in the proof of Theorem 2. The well known RCC-5 algebra is a Constraint Algebra with 5 basic relations: EQ , DR , PO , PP , PP$^{-1}$ . The five relations correspond to the pictorial representation of Figure 3.

If two spatial regions are in one of the relations of MC-4, then only certain relations of RCC-5 can be established between them. Conversely, if certain relations of RCC-5 are established only certain corresponding relations of MC-4 are. This correspondence, however, is not one-to-one. Consider, for instance, a region $x$ and a region $y$, such that $x$ DR $y$, in RCC-5, namely $x$ is disjoint from $y$. Then, each relation of MC-4 can be established between $x$ and $y$. However, if $x$ PP $y$, namely $x$ is a proper part of $y$, then only the CGPP relation can be established between $x$ and $y$. On the other hand, if two regions are congruent, only the relations DR , PO and EQ can exist between $x$ and $y$.

| Rel. of MC-4 | Rel. of RCC-5 |
|---|---|
| CG | {EQ, DR, PO} |
| CGPP | {PP, DR, PO} |
| CGPP$^{-1}$ | {PP$^{-1}$, DR, PO} |
| CNO | {DR, PO} |

Table 2: The basic relations of MC-4 and their counterparts in RCC-5. The meaning of the Table is that when a relation of MC-4 is established, then one of the relations of RCC-5 in second column is established as well.

| Rel. of RCC-5 | Rel. of MC-4 |
|---|---|
| EQ | CG |
| PP | CGPP |
| PP$^{-1}$ | CGPP$^{-1}$ |
| PO | {CG, CGPP, CGPP$^{-1}$, CNO} |
| DR | {CG, CGPP, CGPP$^{-1}$, CNO} |

Table 3: The basic relations of RCC-5 and their counterparts in MC-4.

In Table 3 we set the correspondences between MC-4 basic relations and RCC-5, while in Table 2 we set the correspondences between RCC-5 and MC-4.

The correspondences of the above Tables are not sufficiently strict, to use them in proving that MSAT(MC-4) is an NP-complete problem by a direct polynomial reduction. If we consider a CSP on MC-4, the corresponding RCC-5 CSP is not intractable, since the relations obtained from Table 2 do not define an intractable subset of RCC-5, by means of the complete classification established by Jonsson and Drakengren (1997). Therefore, MSAT(MC-4) is not reducible to RSAT(RCC-5) (the RSAT symbol represents the satisfiability in RCC models) by the corresponding relation mapping of Table 2.

Conversely, we can establish, by means of the above Tables, that among all the possible regions satisfying the relation CG, there exists at least one pair in which the regions are EQ, that when CGPP is established, there exists one pair in a PP relation, and finally that when CNO is established, we have two regions in a PO relation. The summary of this correspondence is reported in Table 4.

This correspondence is obtained from the definition of basic relations. Two regions $a$ and $b$ are congruent *iff* we can roto-translate $a$ by a $T$ in such a way that $T(a)$ EQ $b$ or conversely by $T'$ so that $T'(b)$ EQ $a$. Analogously $a$ is congruent to a part of $b$ *iff* we can roto-translate $a$ by $T$ so that $T(a)$ PP $b$. Finally two regions are in a CNO relation *iff* they can only be in a PO relation or disjoint.

| Rel. of MC-4 | Rel. of RCC-5 |
|---|---|
| CG | EQ |
| CGPP | PP |
| CGPP$^{-1}$ | PP$^{-1}$ |
| CNO | PO |

Table 4: The basic relations of MC-4 and their counterparts in RCC-5 in the special mapping $\Omega$.

The correspondence of Table 4 is called $\Omega$, and a Network of Constraints $\mathcal{G}$ on MC-4 translated in RCC-5 by it is denoted by $\Omega(\mathcal{G})$. The networks which are labelled on all edges by basic relations of MC-4 are henceforth called *scenarios* of MC-4.

Consider a consistent scenario $S$ of MC-4. Applying the composition tables of MC-4 and RCC-5, we clearly derive that the scenario of RCC-5 $\Omega(S)$ is consistent, when the scenario of MC-4 does so. The consequence of above reasoning is the next lemma.

**Lemma 1** *If a scenario on MC-4 is consistent, then $\Omega(\mathcal{G})$ is consistent.*

Ladkin and Maddux (1994) proved that a network of constraints is consistent *iff* it has a consistent scenario. Therefore, an immediate consequence of Lemma 1 is that if a network of constraints $\mathcal{G}$ on MC-4 is consistent, then $\Omega(\mathcal{G})$ is consistent. We are now able to prove a first theorem.

**Theorem 1** MSAT(MC-4) *is NP-hard.*

**Proof**
By the observation about existence of consistent scenarios in a Constraint Algebra due to Mackworth and Freuder (1985), we obtain a polynomial reduction of RSAT(RCC-5) to MSAT(MC-4).

It suffices to note that if we can solve MSAT then we can solve RCC-5 problems obtained by the $\Omega$ translation as well. Now, the problems mapped from MC-4 into RCC-5 by means of $\Omega$ can be trivially inverted by $\Omega^{-1}$, since $\Omega$ is trivially one-to-one. This means that each problem in the set of relations obtained in RCC-5 by $\Omega$ corresponds to a problem in MC-4, and vice versa.

The set of relations of RCC-5 translated by $\Omega$ contains $\{PP, PP^{-1}\}$ and PO . Nebel and Renz (1999), proved that each set of relations of RCC-5 containing these two relations is intractable. Therefore the set $\Omega(MC-4)$ is intractable.

This proves that if we are able to solve a problem in MC-4 we can solve a problem of a subset of RCC-5 which is intractable. Therefore MSAT(MC-4) is NP-hard.

$\square$

Mackworth and Freuder (1985) also proved that backtracking can be applied to CSPs. The backtracking algorithm is usually implemented by a linear non-deterministic technique, being therefore a polynomial algorithm on non-deterministic machines.

The backtracking technique is applicable to MC-4 as well, so we can perform a polynomial solution of MSAT on nondeterministic machines. This shows that MSAT is in NP, and allows us to claim:

**Theorem 2** MSAT(MC-4) *is NP-complete.*

Because of this negative result a deep investigation is needed to define tractable subclasses of the set of 16 relations which allow us to perform polynomial analysis for at least a subset of the networks of constraints definable on MC-4.

In this paper we give the definition of the three maximal tractable subclasses of MC-4, exhibiting therefore a complete classification of tractability for the algebra. The three maximal tractable subclasses have already been studied by Anger, Mitra and Rodriguez in (1999). They obtain maximality of the algebras, and exhibited $O(n^3)$ algorithms. We have three main differences here with respect to their paper:

1. The number of classes we individuated is lower than theirs, because they found four maximal tractable subclasses. They failed to note that one of the four subalgebras is a subset of another one. In Table 10 the subalgebra $M_{88}$ corresponds to the fourth algebra of Anger, Mitra and Rodriguez. This subalgebra is tractable, but not maximal.

2. The algorithms we exhibit are all $O(n^2)$ while Anger, Mitra and Rodriguez exhibited only an $O(n^3)$ algorithm for one of the three maximal subsets.

3. Our classification is complete. Thus any subset of MC-4 which is not a subset of one of the three maximal tractable subalgebras we individuated is intractable. Anger, Mitra and Rodriguez did not find this result, since they did not analyse all the subalgebras of PO-time algebra, as we did for MC-4.

## 4. Maximal Tractable Subclasses of MC-4

Given a subset $S$ of a constraint algebra $A$, we indicate by $\widehat{S}$ the set formed by all the relations of $A$ which can be written as expressions of the algebra involving only elements of $S$ and the operations of composition, intersection and converse of relations. This set is often called the *transitive closure* of $S$ with respect to the operations above. We refer to it as the *closure* of $S$. A set $S$ which coincides with its closure is called a subalgebra.

In the previous section we recall the result on NP-completeness for MC-4. The first important observation on the complexity of subalgebras is that, when a subalgebra $B$ does not contain the empty relation, then a network of constraints on $B$ cannot entail a strict contradiction: then it is consistent. So far, the problem is necessarily polynomially solvable, because it is $O(1)$. This is stated in the next lemma.

**Lemma 2** *Given an algebra $A$, if a subalgebra $B$ of $A$ does not contain the empty relation, then* SAT(B) *is polynomial.*

The main consequence of Lemma 2 is that we can limit ourselves in the analysis of subalgebras in MC-4 to these subalgebras which contain the empty relation.

Moreover, since a network of constraints represents the relations in an implicit way, when an edge of a network is not labelled we should interpret it as representing the universal relation. Therefore, the universal relation should be a member of the subalgebras to which we are interested in. We call algebras which contain both the empty and the universal relations *expressive algebras*.

There are 102 expressive subalgebras, we denominate $M_i$ where $i$ varies between 0 to 101. In Tables 5, 6, 7, 8, 10, 9 of Appendix A, the 102 expressive subalgebras of MC-4 are listed.

**Lemma 3** *Given a subset A of MC-4, A is an expressive subalgebra iff A is one of the subalgebras $M_i$ with i between 0 and 101 .*

**Proof**
Consider a subset $A$ of MC-4. We test the closure of $A$ by a LISP program which computes the closure by composition, intersection and converse of a subset of MC-4, and test the presence of empty and universal relation in $A$ by the obvious membership test. The LISP procedure TRANSITIVE-CLOSURE is listed in Online Appendix 1, which accompanies this article. The test succeeds for the subalgebras $M_i$ with $i$ between 0 and 101 and fails for all the other $2^{16} - 102 = 65434$ subsets of MC-4. The claim is therefore proved.

□

The following technical lemma shows that some of the 102 subalgebras individuated above are NP-hard. There are 20 subalgebras of the 102 which are intractable by Lemma 4. The algebras are presented in Table 5 of Appendix A. The proof of Lemma 5 is a trivial consequence of the proof of Theorem 2.

**Lemma 4** *Given a subalgebra A of MC-4, if A contains the relations* CNO *and* {CGPP, CGPP$^{-1}$} *then* MSAT(A) *is NP-hard.*

In the next three subsections we show that three maximal tractable subclasses of MC-4 can be found, so that the only intractable algebras are the 20 listed in Table 5 of Appendix A.

## 4.1 The CG -complete Subalgebra $M_{72}$

When proving that the class $R^5_{28}$ is tractable, Jonsson and Drakengren (1997) observed that any subalgebra of RCC-5 containing only relations including EQ is tractable. This result applies also to MC-4 with respect to the relation CG, and also to the relations CNO, {CGPP, CGPP$^{-1}$}, {CG, CNO}, {CG, CGPP, CGPP$^{-1}$}.

The only relevant cases are CG , CNO and {CGPP, CGPP$^{-1}$} , since the other two cases are included in two of the former three. An algebra formed by relations which all contain CG and by the relation $\perp$ is tractable, because we can obtain an inconsistency *iff* we explicitly have an edge labelled by $\perp$ in the network. Deciding the consistency is therefore an $O(n^2)$ problem.

ALGORITHM M72-CONSISTENCY

**INPUT:**   A constraint network $\mathcal{T}$ on $M_{72}$

**OUTPUT:**   Yes if $\mathcal{T}$ has a solution formed by spatial regions of $\mathbb{R}^2$, no if not.

1. **For**   each edge in $\mathcal{T}$, $\langle x, y \rangle$
        **If** the label on $\langle x, y \rangle$ is $\perp$ **then** return inconsistency
   **Loop**
2.   Return consistency

Figure 4: Algorithm M72-consistency.

$$M_{72} = \{\perp, \, \mathsf{CG}, \, \{\mathsf{CG}, \, \mathsf{CGPP}\}, \, \{\mathsf{CG}, \, \mathsf{CGPP}^{-1}\}, \, \{\mathsf{CG}, \, \mathsf{CNO}\}, \, \{\mathsf{CG}, \mathsf{CGPP}, \, \mathsf{CGPP}^{-1}\},$$
$$\{\mathsf{CG}, \, \mathsf{CGPP}, \, \mathsf{CNO}\}, \, \{\mathsf{CG}, \, \mathsf{CGPP}^{-1}, \, \mathsf{CNO}\}, \, \top\}.$$

Since no contradiction derives from a relation in $M_{72}$ except for $\perp$, Algorithm $M_{72}$-consistency (see Figure 4) solves the consistency checking problem for a network of constraints on $M_{72}$. Thus we can claim the following theorem:

**Theorem 3** *Algorithm $M_{72}$-consistency correctly decides the consistency of a network of constraints $\mathcal{T}$ on $M_{72}$ in $O(n^2)$ time where $n$ is the number of vertices in $\mathcal{T}$.*

The immediate consequence of Theorem 3 is the following theorem:

**Theorem 4** MSAT($M_{72}$) *is polynomial.*

The subalgebras included in $M_{72}$ are in Table 6 of Appendix A.

The 13 subalgebras of Table 6 are not the only subalgebras which can be theoretically obtained based on the method incorporated in Algorithm $M_{72}$-consistency. The same algorithm can be applied to subalgebras formed only with relations containing a symmetrical relation in MC-4 and the empty relation. Then we can also prove the polynomiality for subalgebras of relations all containing $\mathsf{CNO}$, or all containing $\{\mathsf{CGPP}, \mathsf{CGPP}^{-1}\}$, if such subalgebras exist. The subalgebra formed by relations containing $\mathsf{CNO}$ and the empty relation is $M_{78}$, the subalgebra formed only by relations containing $\{\mathsf{CGPP}, \mathsf{CGPP}^{-1}\}$ is $M_{31}$. In Table 7 and in Table 8 of Appendix A the subalgebras of $M_{78}$ and of $M_{31}$ are shown.

In the next subsection we introduce a maximal tractable subclass which includes $M_{78}$, and in subsection 4.3 we introduce an algebra containing $M_{31}$, so that the proof of tractability for subalgebras in Tables 7 and 8 can be derived from these tables as well. Conversely, the subalgebra $M_{72}$ is neither a subalgebra of $M_{99}$ nor a subalgebra of $M_{81}$, so Theorem 3 is an independent result.

## 4.2 The Maximal Tractable Subclass $M_{99}$

We look for a maximal tractable subalgebra containing all the basic relations. The best candidate, based on Table 5, is $M_{99}$, which is the only algebra formed by more than 13 relations which can be polynomial, because we did not show that it is NP-hard by reducing it to an intractable problem over RCC-5.

$M_{99} = \{\perp,$ CG, CGPP, CGPP$^{-1}$, CNO, {CG, CGPP}, {CG, CGPP$^{-1}$}, {CG, CNO}, {CGPP, CNO}, {CGPP$^{-1}$, CNO}, {CG, CGPP, CNO}, {CG, CGPP$^{-1}$, CNO}, $\top\}$.

Fortunately, we can prove the tractability of $M_{99}$, so it is maximal based on the fact that all algebras containing the basic relations are either intractable by Table 5 or subsets of $M_{99}$.

Consider the subset of $M_{99}$,

$G_{99} = \{\{$CG, CGPP$\}, \{$CG, CNO$\}, \{$CGPP, CGPP$^{-1}$, CNO$\}\}$.

The following claim holds.

**Lemma 5** $\widehat{G_{99}} = M_{99}$.

**Proof**

The following expressions represent valid implementations of relations in $M_{99}$ using only elements of $G_{99}$ and the operators of composition, intersection and converse.

| t.1. | $\perp$ | $= \{$CG, CGPP$\} \oplus \{$CG, CNO$\} \oplus \{$CGPP, CGPP$^{-1}$, CNO$\}$ |
|------|---------|---|
| t.2. | CG | $= \{$CG, CGPP$\} \oplus \{$CG, CGPP$\}^{\smile}$ |
| t.3. | CGPP | $= \{$CG, CGPP$\} \oplus \{$CGPP, CGPP$^{-1}$, CNO$\}$ |
| t.4. | CGPP$^{-1}$ | $= \{$CG, CGPP$\}^{\smile} \oplus \{$CGPP, CGPP$^{-1}$, CNO$\}$ |
| t.5. | CNO | $= \{$CG, CNO$\} \oplus \{$CGPP, CGPP$^{-1}$, CNO$\}$ |
| t.6. | $\top$ | $= \{$CG, CGPP$\} \otimes \{$CGPP, CGPP$^{-1}$, CNO$\}$ |
| t.7. | {CG, CGPP$^{-1}$} | $= \{$CG, CGPP$\}^{\smile}$ |
| t.8. | {CGPP, CNO} | $= (\{$CG, CGPP$\} \otimes \{$CG, CNO$\}) \oplus \{$CGPP, CGPP$^{-1}$, CNO$\}$ |
| t.9. | {CGPP$^{-1}$, CNO} | $= (\{$CG, CGPP$\}^{\smile} \otimes \{$CG, CNO$\}) \oplus \{$CGPP, CGPP$^{-1}$, CNO$\}$ |
| t.10. | {CG, CGPP, CNO} | $= \{$CG, CGPP$\} \otimes \{$CG, CNO$\}$ |
| t.11. | {CG, CGPP$^{-1}$, CNO} | $= \{$CG, CGPP$\}^{\smile} \otimes \{$CG, CNO$\}$ |

$\square$

A network of constraints $\mathcal{T}$ on $M_{99}$, implemented by means of Lemma 5 is denoted henceforth by $\Psi_{99}(\mathcal{T})$.

We can derive a contradiction from a network of constraints *iff* the network derives two relations $R_1$ and $R_2$, between one pair of vertices such that $R_1 \cap R_2 = \varnothing$. The ways in which a contradiction can be derived in networks labelled by relations of $M_{99}$, based on the intersections of relations (except the empty relation which obviously generates a contradiction by itself) are:

| | | | |
|---|---|---|---|
| a) | CG | $\oplus$ | CGPP |
| b) | CG | $\oplus$ | CNO |
| c) | CG | $\oplus$ | {CGPP, CNO} |
| d) | CG | $\oplus$ | {CGPP, CGPP$^{-1}$, CNO} |
| e) | CGPP | $\oplus$ | CGPP$^{-1}$ |
| f) | CGPP | $\oplus$ | CNO |
| g) | CGPP | $\oplus$ | {CG, CGPP$^{-1}$} |
| h) | CGPP | $\oplus$ | {CG, CNO} |
| i) | CGPP | $\oplus$ | {CGPP$^{-1}$, CNO} |
| l) | CGPP | $\oplus$ | {CG, CGPP$^{-1}$, CNO} |
| m) | CNO | $\oplus$ | {CG, CGPP} |
| n) | {CG, CGPP} | $\oplus$ | {CGPP$^{-1}$, CNO} |

This is simply obtained by considering all pairs of relations in $M_{99}$ whose intersection is empty.

In $G_{99}$ the contradictions are only:
{CG, CGPP} $\oplus$ {CG, CGPP} $^{\smile}$ $\oplus$ {CGPP, CGPP$^{-1}$, CNO} and
{CG, CGPP} $\oplus$ {CG, CNO} $\oplus$ {CGPP, CGPP$^{-1}$, CNO} .
Henceforth we represent the relation {CG, CGPP} by $\precsim$, the relation {CG, CNO} by $\overset{\bowtie}{\sim}$ and the relation {CGPP, CGPP$^{-1}$, CNO} by $\not\precsim$. Since any path in which labels are all $\precsim$ corresponds to the representation of $\precsim$ between each pair of vertices in the path, and $\precsim \otimes \overset{\bowtie}{\sim} = \precsim$, the possible expressions for contradictions in $G_{99}$ are $(\precsim^n \otimes \precsim^{-n}) \oplus \not\precsim$ and $(\precsim^n \otimes \precsim^{-(n-k)} \otimes \overset{\bowtie}{\sim} \otimes \precsim^{-k}) \oplus \not\precsim$. Hence, a cycle $(\precsim^n \otimes \precsim^{-n})$ is called a $\precsim$-cycle, while a cycle $(\precsim^n \otimes \precsim^{-(n-k)} \otimes \overset{\bowtie}{\sim} \otimes \precsim^{-k})$ is called a quasi $\precsim$-cycle. The graph representation of these two different contradictory situations in $G_{99}$ is showed in Figure 5. A $\precsim$-cycle and a quasi $\precsim$-cycle, in $M_{99}$ force the elements involved in the cycle to be all in the relation CG .

We can now show that the contradictions represented in $G_{99}$ are the only contradictions which can be obtained by the implementation suggested in Lemma 5. This is very important, because we may perform consistency checking by simply checking all cycles. This result is stated in the next lemma.

**Lemma 6** *Given a network of constraints $\mathcal{T}$ on $M_{99}$, $\mathcal{T}$ is inconsistent iff $\Psi_{99}(\mathcal{T})$ contains either a $\precsim$-cycle of a quasi $\precsim$-cycle, and two vertices of one cycle are connected by an edge labelled by $\not\precsim$.*

**Proof (Sketch)**

Case by case, the contradictions from a) to n), as in the table above, generate one of the two situations of the claim.

For example, the contradiction a) CG ⊕ CGPP , generates

$$(\{CG, CGPP\} \otimes \{CG, CGPP\}^{\smile}) \oplus (\{CG, CGPP\} \oplus \{CGPP, CGPP^{-1}, CNO\})$$

which is a $\precsim$-cycle in which two vertices are connected by an edge labelled by $\not\prec$, as stated in the claim. All the other cases behave in the same way as can be easily checked by the reader.

Conversely, if a contradiction derives from one of the possible implicit ways of representing relations in $M_{99}$, the implementation also produces one of the cases of the claim. In particular we have that, in $M_{99}$, $CG^n = CG$ , $CGPP^n = CGPP$ , $(CGPP^{-1})^n = CGPP^{-1}$ , $\{CG, CGPP\}^n = \{CG, CGPP\}$ and $\{CG, CGPP^{-1}\}^n = \{CG, CGPP^{-1}\}$ are the only idempotent relations. The other cases of implicitness can be obtained by considering all the $14 \cdot 14 = 196$ pairs of relations in $M_{99}$, composing and intersecting them. The implicit cases arising thus are listed in below.

| | | |
|---|---|---|
| i.1. | CG | $= \{CG, CGPP\}^n \oplus \{CG, CGPP^{-1}\}^n$ |
| i.2. | CG | $= \{CG, CGPP\}^n \oplus \{CG, CNO\}$ |
| i.3. | CGPP | $= \{CG, CGPP\}^n \oplus \{CGPP, CNO\}$ |
| i.4. | CNO | $= \{CG, CNO\} \oplus \{CGPP, CNO\}$ |
| i.5. | CNO | $= \{CGPP, CNO\} \oplus \{CGPP^{-1}, CNO\}$ |
| i.6. | CNO | $= \{CGPP, CNO\} \oplus \{CG, CGPP^{-1}, CNO\}$ |
| i.7. | $\{CG, CNO\}$ | $= \{CG, CGPP, CNO\} \qquad \oplus \{CG, CGPP^{-1}, CNO\}$ |
| i.8. | $\{CGPP, CNO\}$ | $= CGPP^n \otimes CNO$ |
| i.9. | $\{CGPP, CNO\}$ | $= CGPP^n \otimes \{CG, CNO\}$ |
| i.10. | $\{CGPP, CNO\}$ | $= CGPP^n \otimes \{CG, CGPP, CNO\}$ |
| i.11. | $\{CGPP, CNO\}$ | $= \{CG, CGPP\}^n \otimes CNO$ |
| i.12. | $\{CGPP, CNO\}$ | $= \{CG, CGPP, CNO\} \qquad \oplus \{CGPP, CGPP^{-1}, CNO\}$ |
| i.13. | $\{CG, CGPP, CNO\}$ | $= \{CG, CGPP\}^n \otimes \{CG, CNO\}$ |

Readers may directly express in $G_{99}$ each of these relations along with the corresponding relation producing a contradiction in $M_{99}$ and they immediately verify the existence of $\precsim$-cycles or quasi $\precsim$-cycle where one pair of vertices is connected by an edge labelled by $\not\prec$ for these generated graphs.

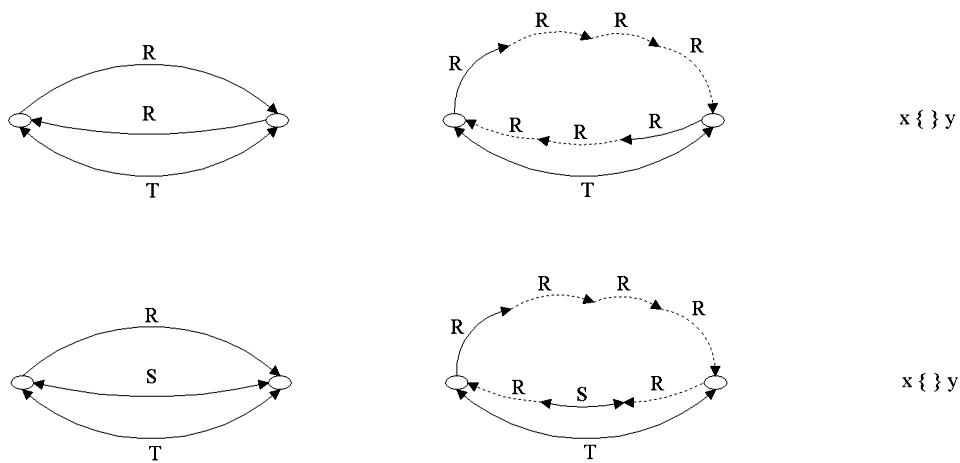

Figure 5: Contradictions in $G_{99}$: {} indicates $\perp$ and R, S, T respectively {CG, CGPP}, {CG, CNO} and {CGPP, CGPP$^{-1}$, CNO}.

For example if we express CG as implicit relation by i.1. as in table above (formed by relations in $G_{99}$), CGPP as implicit relation by i.3. and {CGPP, CNO} by $\Psi_{99}$ as in t.8., and consider the contradiction CG $\oplus$ CGPP we obtain a quasi $\precsim$-cycle where two vertices are connected by an edge labelled by $\not\prec$. Similarly we can derive the other cases.

$\square$

We can now exhibit an algorithm which looks for $\precsim$-cycles and quasi $\precsim$-cycles, and checks about pairs of a cycle being connected by an edge labelled by $\not\prec$.

In Figure 7 an algorithm able to solve the Consistency Checking Problem for networks of constraints labelled by relations in $M_{99}$ is presented. Based on Lemma 6 we can prove the following theorem.

**Theorem 5** *Algorithm M99-CONSISTENCY correctly decides the consistency of a network of constraints $\mathcal{T}$ on $M_{99}$ in $O(n^2)$ steps where n is the number of vertices of $\mathcal{T}$.*

**Proof**
By Lemma 6 we can ensure the correctness of Algorithm M99-CONSISTENCY. The complexity of the algorithm can be derived from the fact that the computation of strongly

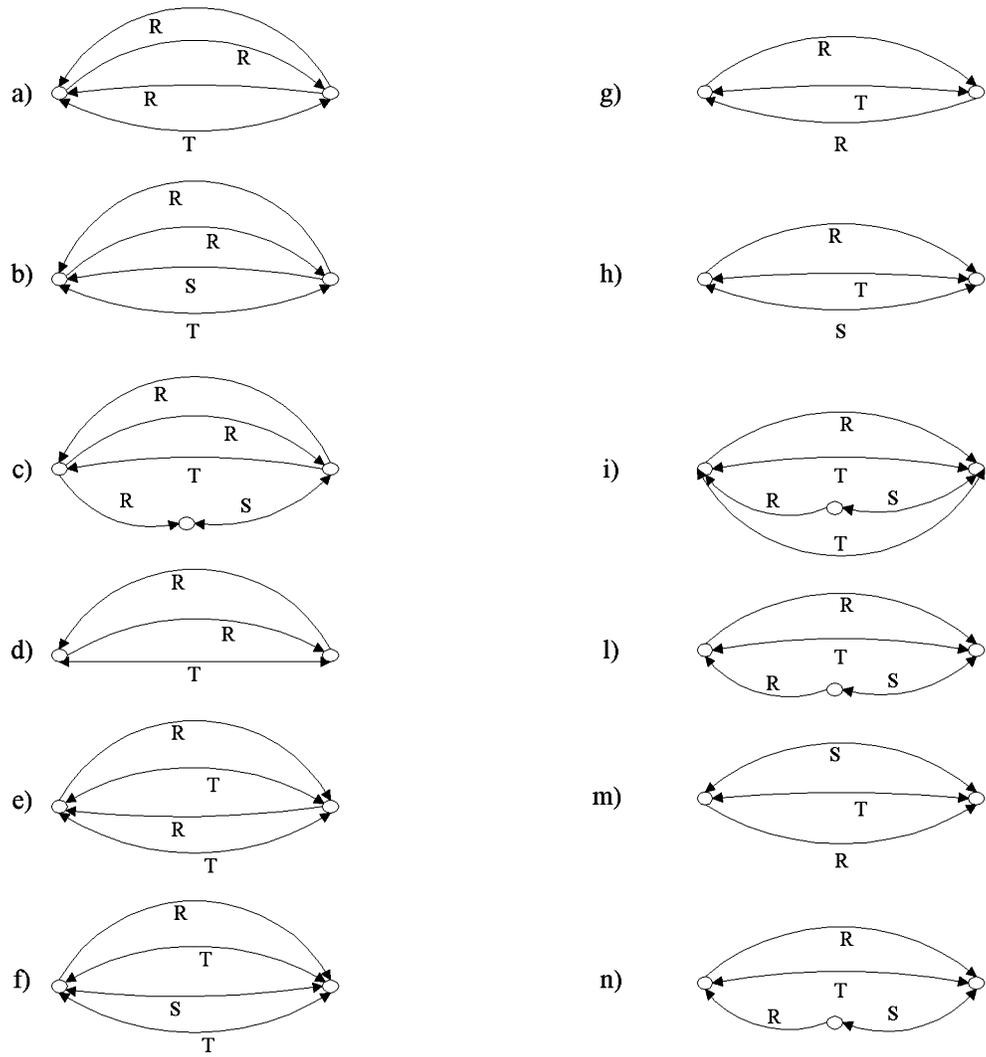

Figure 6: Contradictions in $M_{99}$: R, S, T respectively indicate {CG, CGPP}, {CG, CNO} and {CGPP, CGPP$^{-1}$, CNO}. Letters from a) to n) refer to the table of page 375.

ALGORITHM M99-CONSISTENCY

**INPUT:**    A constraint network $\mathcal{T}$ on $M_{99}$

**OUTPUT:**   Yes if $\mathcal{T}$ has a solution formed by spatial regions of $\mathbb{R}^2$, no if not.

1.    **Translate** $\mathcal{T}$ into the generator set as in Lemma 5
2.    **Look** for cycles labelled by $\precsim$ and at most one $\bowtie$
3.    **Check** whether edges between two vertices in one cycle are not labelled by $\not\sim$ otherwise **return** *no*.
4.    **Substitute** vertices of the cycle with one vertex.
2.    **If** there are more cycles **go to** Step 2., otherwise **return** *yes*.

Figure 7: Algorithm M99-consistency

connected components is a $O(e)$ problem where $e$ is the number of edges in the network. The $\Psi_{99}$ implementation adds, in the worst case, $O(e')$ edges, where $e'$ is the number of edges in $\mathcal{T}$, and therefore the number of edges in $\Psi_{99}(\mathcal{T})$ is $O(2 \cdot e')$, which is $O(n^2)$.

$\square$

An immediate consequence of Theorem 5 is the following theorem:

**Theorem 6** MSAT($M_{99}$) *is polynomial.*

The subalgebras of MC-4 included in $M_{99}$ are in Table 10 of Appendix A.

### 4.3 The Maximal Tractable Subclass $M_{81}$

The problem of deciding the consistency of a network of constraints on MC-4 is tractable, by means of Tables 6, 7, 8 and 10 for 85 subalgebras. The remainder is formed by 17 subalgebras, each of these can be either tractable or intractable. The set

$$M_{81} = \{ \perp, \text{CG, CGPP, CGPP}^{-1}, \{\text{CG, CGPP}\}, \{\text{CG, CGPP}^{-1}\}, \{\text{CGPP, CGPP}^{-1}\}, \{\text{CG, CGPP, CGPP}^{-1}\}, \{\text{CGPP, CGPP}^{-1}, \text{CNO}\}, \top\}$$

contains all these subalgebras, so that if $M_{81}$ is tractable all these algebras are as well.

By analogy with the schema of the previous section we look for a small generator set for $M_{81}$. This is $G_{81} = \{\{\text{CG, CGPP}\}, \{\text{CG, CGPP, CGPP}^{-1}\}, \{\text{CGPP, CGPP}^{-1}, \text{CNO}\}\}$. The following lemma states the properties of $M_{81}$ with respect to $G_{81}$.

**Lemma 7** $\widehat{G_{81}} = M_{81}$.

**Proof**

The following expressions represent valid implementations of relations in $M_{81}$ using only elements of $G_{81}$ and the operators of composition, intersection and converse.

| r.1. | $\perp$ | = | $\{CG, CGPP\} \quad \oplus \quad \{CG, CGPP\}^{\smile}$ <br> $\oplus \{CGPP, CGPP^{-1}, CNO\}$ |
|------|---------|---|---|
| r.2. | CG | = | $\{CG, CGPP\} \oplus \{CG, CGPP\}^{\smile}$ |
| r.3. | CGPP | = | $\{CG, CGPP\} \oplus \{CGPP, CGPP^{-1}, CNO\}$ |
| r.4. | $CGPP^{-1}$ | = | $\{CG, CGPP\}^{\smile} \oplus \{CGPP, CGPP^{-1}, CNO\}$ |
| r.5. | $\{CG, CGPP^{-1}\}$ | = | $\{CG, CGPP\}^{\smile}$ |
| r.6. | $\{CGPP, CGPP^{-1}\}$ | = | $\{CG, CGPP, CGPP^{-1}\} \qquad \oplus$ <br> $\{CGPP, CGPP^{-1}, CNO\}$ |
| r.7. | $\top$ | = | $\{CG, CGPP\} \otimes \{CGPP, CGPP^{-1}, CNO\}$ |

$\square$

The implementation of relations in $M_{81}$ as described in Lemma 7 is denoted by $\Psi_{81}$. The contradictions in $M_{81}$ are as in the next table.

| a) | CG | $\oplus$ | CGPP |
|----|----|----------|------|
| b) | CG | $\oplus$ | $\{CGPP, CGPP^{-1}\}$ |
| c) | CG | $\oplus$ | $\{CGPP, CGPP^{-1}, CNO\}$ |
| d) | CGPP | $\oplus$ | $CGPP^{-1}$ |
| e) | CGPP | $\oplus$ | $\{CG, CGPP^{-1}\}$ |

The only possible contradiction in $G_{81}$ is

$$\{CG, CGPP\} \oplus \{CG, CGPP\}^{\smile} \oplus \{CGPP, CGPP^{-1}, CNO\}$$

and it corresponds to a $\precsim$-cycle where two vertices are connected by an edge labelled by $\not\prec$.

**Lemma 8** *Given a network of constraints $\mathcal{T}$ on $M_{81}$, $\mathcal{T}$ is inconsistent iff $\Psi_{81}(\mathcal{T})$ contains a $\precsim$-cycle and two vertices of one cycle are connected by an edge labelled by $\not\prec$.*

**Proof (Sketch)**
Case by case, the contradictions from a) to e), as in the table above, generate one of the two cases claimed here.

For example, the contradiction $CG \oplus \{CGPP, CGPP^{-1}\}$ , is implemented

$$(\{CG, CGPP\} \oplus \{CG, CGPP\}^{\smile}) \oplus (\{CG, CGPP, CGPP^{-1}\}$$
$$\oplus \{CGPP, CGPP^{-1}, CNO\})$$

which is a $\precsim$-cycle in which two vertices are connected by an edge labelled by $\not\prec$, as stated in the claim. All the other cases behave in the same way as can be easily checked by the reader.

The only possible ways of representing implicit relations are provided by the schema of $\Psi_{81}$. Therefore the claim is proved.

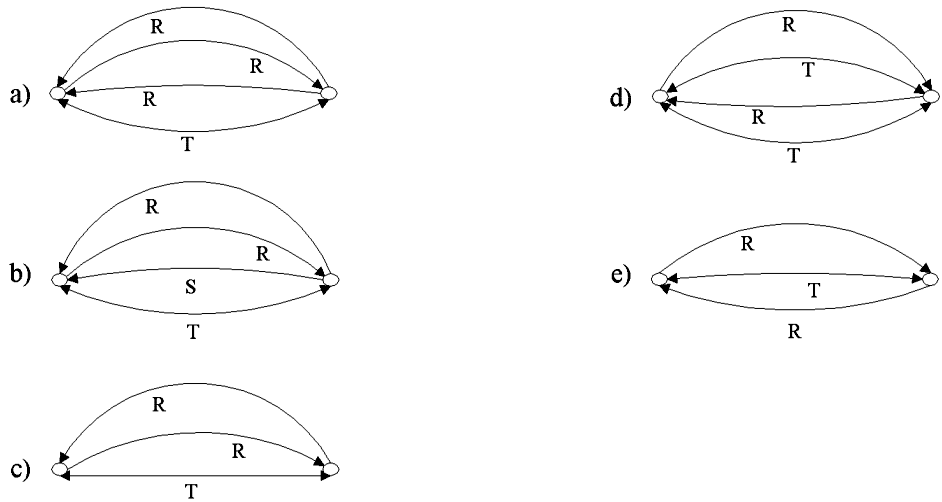

Figure 8: Contradictions in $M_{81}$. R, S, T respectively indicate {CG, CGPP}, {CG, CGPP, CGPP$^{-1}$}, {CGPP, CGPP$^{-1}$, CNO}. Letters from a) to e) refer to the table of page 380.

## ALGORITHM M81-CONSISTENCY

**INPUT**:    A constraint network $\mathcal{T}$ on $M_{81}$

**OUTPUT**:    Yes if $\mathcal{T}$ has a solution formed by spatial regions of $\mathbb{R}^2$, no if not.

1.    **Translate** $\mathcal{T}$ into the generator set as in Lemma 7
2.    **Look** for cycles labelled by $\precsim$
3.    **Check** whether edges between two vertices in one cycle are not labelled by $\not\prec$ otherwise **return** *no*.
4.    **Substitute** vertices of the cycle with one vertex.
2.    **If** there are more cycles **go to** Step 2., otherwise **return** *yes*.

Figure 9: Algorithm M81-consistency

□

In Figure 9 an algorithm is presented which solves the problem of consistency checking for the subalgebra $M_{81}$. We can show, in particular, the following claim.

**Theorem 7** *Algorithm M81-CONSISTENCY correctly decides the consistency of a network of constraints $\mathcal{T}$ on $M_{81}$ in $O(n^2)$ steps where $n$ is the number of vertices of $\mathcal{T}$.*

**Proof**
By Lemma 8 we can ensure the correctness of Algorithm M81-CONSISTENCY. The complexity of the algorithm can be derived from the fact that the computation of strongly connected components is a $O(e)$ problem where $e$ is the number of edges in the network. The $\Psi_{81}$ implementation adds, in the worst case, $O(e')$ edges, where $e'$ is the number of edges in $\mathcal{T}$, and therefore the number of edges in $\Psi_{81}(\mathcal{T})$ is $O(2 \cdot e')$, which is clearly $O(n^2)$.

□

An immediate consequence of Theorem 3 is the following

**Theorem 8** $\mathrm{MSAT}(M_{81})$ *is polynomial.*

The subalgebras included in $M_{81}$ are in Table 9 of Appendix A.

## 5. Conclusions

We presented a classification of tractability which is complete for the spatial algebra MC-4. This classification states that there exist three maximal tractable subalgebras $M_{72}$, $M_{99}$ and $M_{81}$ which include 92 out of 102 expressive subalgebras of MC-4.

The interest in a complete classifications of tractability, as already observed by Jonsson and Drakengren (1997), is determined by the need for a definition of the boundary between tractable and intractable problems. Nebl (1999) has suggested that the knowledge of this boundary can be used either as preprocessing step and as a way to structure the backtracking search on such algebras.

The provision of a complete classification is one step in researching about constraint algebras. A next step is the individuation of useful heuristics which give improvements in the performances of various techniques. We are currently exploiting the use of these techniques in association with techniques based on the classification presented in this paper to obtain efficient reasoning algorithms which can be used in practice. Preliminary results on path-consistency are encouraging, but we cannot yet guarantee the percentage of improvement, since the algebra MC-4 is so simply structured that for networks randomly chosen it is very hard to obtain a case where inconsistency is not detectable by path-consistency.

## Acknowledgements

I would like to thank Elena Mutinelli who developed her Laurea Thesis (Mutinelli, 1998) on the theme of complexity of reasoning about congruence and first discussed preliminary results which I used for developing the classification of this paper. Her work has been very relevant in reaching my results.

I would also like to thank Bernhard Nebel for some important observations he made to a student of mine, Alessandro Fin, which improved this work. Further thanks go to Jochen Renz for discussions in the early stages of this work.

I would also like to thank the anonymous referees of the Journal of Artificial Intelligence Research for their very careful comments and suggestions which allowed me to make the presentation simpler and more systematical. Some of the proofs in the paper have been rewritten thanks to their suggestions.

Finally I would like to thank Tony Cohn for reading a near final version. His comments have been useful in enhancing both scientific and literary quality of the paper.

This work has been developed in the context of the National Project, MURST ex 40% "Metodologie e tecnologie per la gestione di dati e processi su reti internet ed intranet" (Methods and technologies for data and process management on internet and intranet) directed by L. Tanca.

# Appendix A. Tables of the 102 subalgebras of MC-4

In this section we present the subalgebras of MC-4 organized in separated tables depending on their characteristics. In particular Table 5 shows those subalgebras which are intractable by Lemma 4, Table 6 those which are tractable by Theorem 3, and in Tables 7, 8 those to which Algorithm used in Theorem 3 can be applied (which are all including CNO or {CGPP, CGPP⁻¹} ). In Table 10 the subalgebras included in $M_{99}$ are displyed, while in Table 9 we present the subalgebras included in $M_{81}$.

| Rel. | | | | | | | | | | | | | | | | |
|---|---|---|---|---|---|---|---|---|---|---|---|---|---|---|---|---|
| CG | | × | | | | × | × | × | | | | × | × | × | | × |
| CGPP | | | × | | | × | | | × | × | | × | × | | × | × |
| CGPP⁻¹ | | | | × | | | × | | × | | × | | | × | × | × |
| CNO | | | | | × | | | × | | × | × | | × | × | × | × |
| Alg. | | | | | | | | | | | | | | | | |
| $M_{30}$ | | • | | | | • | | | • | | | | | | • | • |
| $M_{43}$ | | • | • | | | • | | | • | | | | | | • | • |
| $M_{44}$ | | • | | | | • | | • | • | | | | | | • | • |
| $M_{46}$ | | • | | | | • | | | • | | | • | | | • | • |
| $M_{54}$ | | • | • | | | • | | • | • | | | • | | | • | • |
| $M_{56}$ | | • | • | | | • | | | • | | | | | | • | • |
| $M_{67}$ | | • | • | | | • | | • | • | | | • | | | • | • |
| $M_{77}$ | | • | | • | • | • | | | • | • | • | | | | • | • |
| $M_{83}$ | | • | • | • | • | • | | | • | • | • | | | | • | • |
| $M_{84}$ | | • | | • | • | • | | • | • | • | | | | | • | • |
| $M_{85}$ | | • | | • | • | • | | | • | • | • | • | | | • | • |
| $M_{89}$ | | • | • | • | • | • | | • | • | • | | | | | • | • |
| $M_{90}$ | | • | • | • | • | • | | | • | • | • | • | | | • | • |
| $M_{92}$ | | • | • | • | • | • | • | • | • | • | • | | | | • | • |
| $M_{93}$ | | • | • | • | • | • | | | • | • | • | • | • | | • | • |
| $M_{95}$ | | • | | • | • | • | | • | • | • | • | • | | • | • | • |
| $M_{97}$ | | • | • | • | • | • | • | • | • | • | • | • | | | • | • |
| $M_{98}$ | | • | • | • | • | • | | | • | • | • | • | | • | • | • |
| $M_{100}$ | | • | • | • | • | • | • | • | • | • | • | • | | • | • | • |
| $M_{101}$ | | • | • | • | • | • | • | • | • | • | • | • | • | • | • | • |

Table 5: The subalgebras of MC-4 containing the relations CNO and {CGPP, CGPP⁻¹}

| Rel. | CG | | × | | | | × | × | × | | | | × | × | × | | × |
|------|-----|---|---|---|---|---|---|---|---|---|---|---|---|---|---|---|---|
| | CGPP | | | × | | | × | | | | × | × | × | × | | × | × |
| | CGPP⁻¹ | | | | × | | | × | | × | | × | | × | × | | × | × |
| | CNO | | | | | × | | | × | | | × | × | | × | × | × | × |
| Alg. | | | | | | | | | | | | | | | | | |
| $M_0$ | | | • | | | | | | | | | | | | | | • |
| $M_1$ | | | • | • | | | | | | | | | | | | | • |
| $M_3$ | | | • | | | | | | × | | | | | | | | • |
| $M_5$ | | | • | | | | | | | | | | • | | | | • |
| $M_9$ | | | • | • | | | | | • | | | | | | | | • |
| $M_{12}$ | | | • | • | | | | | | | | | • | | | | • |
| $M_{18}$ | | | • | • | | | | • | • | | | | | | | | • |
| $M_{22}$ | | | • | • | | | | | • | | | | • | | | | • |
| $M_{25}$ | | | • | | | | | | • | | | | | • | • | | • |
| $M_{34}$ | | | • | • | | | • | • | | | | | • | | | | • |
| $M_{38}$ | | | • | • | | | | | • | | | | • | | | | • |
| $M_{63}$ | | | • | • | | | • | • | • | | | | | • | • | | • |
| $M_{72}$ | | | • | • | | | • | • | • | | | | • | • | • | | • |

Table 6: The subalgebras of MC-4 contained in $M_{72}$.

| Rel. | CG | | × | | | | × | × | × | | | | × | × | × | | × |
|------|-----|---|---|---|---|---|---|---|---|---|---|---|---|---|---|---|---|
| | CGPP | | | × | | | × | | | | × | × | | × | × | | × | × |
| | CGPP⁻¹ | | | | × | | | × | | × | | × | | × | × | | × | × |
| | CNO | | | | | × | | | × | | | × | × | | × | × | × | × |
| Alg. | | | | | | | | | | | | | | | | | |
| $M_2$ | | | • | | | • | | | | | | | | | | | • |
| $M_6$ | | | • | | | | | | | | | | | | × | | • |
| $M_{10}$ | | | • | | | • | | | • | | | | | | | | • |
| $M_{15}$ | | | • | | | • | | | | | | | | | × | | • |
| $M_{24}$ | | | • | | | • | | | • | • | | | | | | | • |
| $M_{28}$ | | | • | | | • | | • | | | | | | | × | | • |
| $M_{37}$ | | | • | | | • | | • | • | • | | | | | | | • |
| $M_{39}$ | | | • | | | • | | • | | | | | | • | • | | • |
| $M_{47}$ | | | • | | | • | | | • | • | | | | | × | | • |
| $M_{58}$ | | | • | | | • | | • | • | • | | | | | × | | • |
| $M_{64}$ | | | • | | | • | | • | • | • | | | | • | • | | • |
| $M_{78}$ | | | • | | | • | | • | • | • | • | | | • | • | • | • |

Table 7: The subalgebras of MC-4 contained in $M_{78}$ and not contained in $M_{72}$.

| Rel. | CG | | × | | | | × | × | × | | | | × | × | × | | × |
|---|---|---|---|---|---|---|---|---|---|---|---|---|---|---|---|---|---|
| | CGPP | | | × | | | | × | | | × | × | × | × | | × | × |
| | CGPP⁻¹ | | | | × | | | | × | | | × | | × | × | × | × |
| | CNO | | | | | × | | | | × | | | × | × | × | × | × |
| Alg. | | | | | | | | | | | | | | | | | |
| | $M_4$ | | • | | | | | • | | | | | | | | | • |
| | $M_{13}$ | | • | | | | | • | | | | • | | | | | • |
| | $M_{16}$ | | • | | | | | • | | | | | | | | • | • |
| | $M_{31}$ | | • | | | | | • | | | | • | | | | • | • |

Table 8: The subalgebras of MC-4 contained in $M_{31}$ and not contained in $M_{72}$ or $M_{78}$.

| Rel. | CG | | × | | | | × | × | × | | | | × | × | × | | × |
|---|---|---|---|---|---|---|---|---|---|---|---|---|---|---|---|---|---|
| | CGPP | | | × | | | | × | | | × | × | × | × | | × | × |
| | CGPP⁻¹ | | | | × | | | | × | | | × | | × | × | × | × |
| | CNO | | | | | × | | | | × | | | × | × | × | × | × |
| Alg. | | | | | | | | | | | | | | | | | |
| | $M_{11}$ | | • | • | | | | | | • | | | | | | | • |
| | $M_{20}$ | | • | | • | • | | | | | | | | | | | • |
| | $M_{21}$ | | • | | • | • | | | | | | | | • | | | • |
| | $M_{23}$ | | • | • | | | | | | • | | | | • | | | • |
| | $M_{29}$ | | • | • | | | | | | • | | | | | | • | • |
| | $M_{32}$ | | • | • | • | • | | | | • | | | | | | | • |
| | $M_{33}$ | | • | • | • | • | | | | | | | | • | | | • |
| | $M_{35}$ | | • | | • | • | | | | • | | | | • | | | • |
| | $M_{42}$ | | • | | • | • | | | | • | | | | | | • | • |
| | $M_{45}$ | | • | • | | | | | | • | | | | • | | • | • |
| | $M_{55}$ | | • | | • | • | | | | • | | | | • | | • | • |
| | $M_{59}$ | | • | • | • | • | | • | • | • | | | | | | | • |
| | $M_{60}$ | | • | • | • | • | | • | • | | | | | • | | | • |
| | $M_{66}$ | | • | • | • | • | | | | • | | | | • | | • | • |
| | $M_{70}$ | | • | • | • | • | | • | • | | | | | • | | | • |
| | $M_{74}$ | | • | • | • | • | | • | • | • | | | | | | • | • |
| | $M_{81}$ | | • | • | • | • | | • | • | • | | | | • | | • | • |

Table 9: The subalgebras of MC-4 contained in $M_{81}$ and not contained in $M_{99}$ or $M_{72}$ or $M_{78}$ or $M_{31}$.

| Rel. | CG | | × | | | | × | × | × | | | | × | × | × | | × |
|------|-----|---|---|---|---|---|---|---|---|---|---|---|---|---|---|---|---|
| | CGPP | | | × | | | × | | | × | × | | × | × | | | × | × |
| | CGPP$^{-1}$ | | | | × | | | × | | × | | × | × | | × | × | × | × |
| | CNO | | | | | × | | | × | | × | × | | × | × | × | × | × |
| Alg. | | | | | | | | | | | | | | | | | |
| $M_7$ | | | • | | • | • | | | | | | | | | | | • |
| $M_8$ | | | • | • | | • | | | | | | | | | | | • |
| $M_{14}$ | | | • | • | | | | | | | | | | | • | • | • |
| $M_{17}$ | | | • | • | • | • | | | | | | | | | | | • |
| $M_{19}$ | | | • | • | | | • | | | • | | | | | | | • |
| $M_{26}$ | | | • | | • | • | | | | | | | | | • | • | • |
| $M_{27}$ | | | • | • | | | • | | | | | | | | | | • |
| $M_{36}$ | | | • | • | | | • | | | | | | • | • | | | • |
| $M_{40}$ | | | • | • | • | • | | | | | | | | | • | • | • |
| $M_{41}$ | | | • | • | | | • | | | • | | | | | • | • | • |
| $M_{48}$ | | | • | • | • | • | | • | • | | | | | | | | • |
| $M_{50}$ | | | • | | • | • | • | | | | | | • | • | | | • |
| $M_{51}$ | | | • | • | | | • | | | • | | | • | • | | | • |
| $M_{52}$ | | | • | • | | | • | | | • | | | | | • | • | • |
| $M_{57}$ | | | • | • | | | • | | | | | | • | • | | • | • |
| $M_{61}$ | | | • | • | • | • | • | | | | | | • | • | | | • |
| $M_{62}$ | | | • | | • | • | • | | | • | | | • | • | | | • |
| $M_{65}$ | | | • | • | • | | • | • | • | | | | | | • | • | • |
| $M_{68}$ | | | • | | • | • | • | | | | | | • | • | | • | • |
| $M_{69}$ | | | • | • | | | • | | | • | | | • | • | | • | • |
| $M_{71}$ | | | • | • | • | • | • | | | • | | | | | | | • |
| $M_{73}$ | | | • | • | | | • | | | • | | | • | • | | • | • |
| $M_{75}$ | | | • | • | • | | • | | | | | | • | • | | • | • |
| $M_{76}$ | | | • | | • | • | • | | | • | | | • | • | | • | • |
| $M_{79}$ | | | • | • | • | • | • | • | • | | | | • | • | | | • |
| $M_{80}$ | | | • | | • | • | • | | | • | | | • | • | | • | • |
| $M_{82}$ | | | • | • | • | • | • | | | • | | | • | • | | | • |
| $M_{86}$ | | | • | • | | | • | | | • | | | • | • | | • | • |
| $M_{87}$ | | | • | • | • | • | • | | | • | | | • | • | | • | • |
| $M_{88}$ | | | • | • | • | • | • | • | • | | | | • | • | | • | • |
| $M_{91}$ | | | • | | • | • | • | | | • | | | • | • | | • | • |
| $M_{94}$ | | | • | • | • | • | • | | | • | | | • | • | | • | • |
| $M_{96}$ | | | • | • | • | • | • | • | • | • | | | • | • | | • | • |
| $M_{99}$ | | | • | • | • | • | • | • | • | • | • | | • | • | | • | • |

Table 10: The subalgebras of MC-4 contained in $M_{99}$ and not contained in $M_{72}$ or $M_{78}$ or $M_{31}$.